\documentclass[preprint,12pt]{elsarticle}

\usepackage{amssymb}
\usepackage{amsmath}
\usepackage{booktabs}
\renewcommand{\vec}[1]{\boldsymbol{#1}} 

\journal{ENGINEERING APPLICATIONS OF ARTIFICIAL INTELLIGENCE}

\begin{document}

\begin{frontmatter}




\title{BERT-based Chinese Text Classification for Emergency Domain with a Novel Loss Function}


\author[label1,label2]{Zhongju Wang}
\author[label1,label2]{Long Wang\corref{mycorrespondingauthor}}
\ead{Long.Wang@IEEE.org}
\cortext[mycorrespondingauthor]{Corresponding author}
\author[label1,label2]{Chao Huang}
\author[label1,label2]{Xiong Luo}

\address[label1]{School of Computer and Communication Engineering, University of Science and Technology Beijing, Beijing 100083, China}
\address[label2]{Beijing Key Laboratory of Knowledge Engineering for Materials Science, Beijing 100083, China}


\begin{abstract}

This paper proposes an automatic Chinese text categorization method for solving the emergency event report classification problem. Since bidirectional encoder representations from transformers (BERT) has achieved great success in natural language processing domain, it is employed to derive emergency text features in this study. To overcome the data imbalance problem in the distribution of emergency event categories, a novel loss function is proposed to improve the performance of the BERT-based model. Meanwhile, to avoid the impact of the extreme learning rate, the Adabound optimization algorithm that achieves a gradual smooth transition from Adam to SGD is employed to learn parameters of the model. To verify the feasibility and effectiveness of the proposed method, a Chinese emergency text dataset collected from the Internet is employed. Compared with benchmarking methods, the proposed method has achieved the best performance in terms of accuracy, weighted-precision, weighted-recall, and weighted-F1 values. Therefore, it is promising to employ the proposed method for real applications in smart emergency management systems.

\end{abstract}


\end{frontmatter}


\section{Introduction}
\label{}

Public emergency events and natural disasters occur frequently in last few decades. As a result, it has brought serious negative effects on the economy and society \cite{2019Agent}. Hence, disaster prevention and mitigation have become an urgent task to maintain economic development and social stability. In response, governments have designed corresponding emergency management systems to better handle these disasters \cite{de2019creative}. Considering the diversity and high frequency of emergency events, rapid emergency response is a serious challenge for current emergency management systems. Intelligent emergency classification is conducive to the quick decision of disposal plan and reasonable allocation of resources for improving emergency preparedness \cite{2020Improving}.

In China, with the development of information technologies, thousands of emergency events are reported in the Internet every year. If these text data can be classified and processed to form a standardized emergency database, it can not only be used for public safety education but also for providing reference solutions for emergency decision-making. However, due to the characteristics of Chinese text, such as unorganized structure and huge information content, it takes a lot of time and effort to classify and organize these emergency events manually. Therefore, emergency event text classification is a challenging task in the emergency field of China.

Traditional machine learning methods for text classification are based on statistical learning, such as Bayesian classifier,  K-Nearest Neighbors (KNN), support vector machines (SVM), decision tree. Main problems of such methods are that these lack the capability of feature extraction since the text representation is high-dimensional and highly sparse. In addition, conventional machine learning methods require feature engineering, which is very time-consuming. In recent years, deep learning has achieved great success in the field of natural language processing, such as extractive summarization \cite{2018Deep}, machine translation \cite{singh2017machine}, and text classification \cite{semberecki2017deep}. Different from traditional machine learning methods, deep learning models can automatically extract text features without human labor involvement, which provides a foundation for utilizing deep learning to process public emergency event report. However, there are still many challenges when apply deep learning models to classifying emergency events. On the one hand, long emergency event text lead the model to lose certain semantic information. On the other hand, due to text data is generally unstructured and contains complex semantic relationships, it is difficult to train a good model with a small imbalanced corpus. In this paper, the current state-of-the-art deep learning-based natural language processing model, Bidirectional Encoder Representations from Transformers (BERT) \cite{devlin2018bert}, is employed to derive emergency text features. To overcome the data imbalance problem, we propose a novel loss function to improve the classification accuracy of the BERT-based model. The main contributions of this study are summarized as follows:

\begin{itemize}
	\item [(1)]
	A novel loss function is proposed to improve the performance of the BERT-based classifier on imbalanced data for the emergency domain.
	\item [(2)]
	To guarantee the convergence speed and model training quality, the Adabound optimizer that achieved a gradual smooth transition from Adam to SGD is employed to learn the model parameters.
	\item [(3)]
	The feasibility and effectiveness of the proposed method are validated based on a real Chinese emergency text dataset collected from the Internet. 
\end{itemize}

The rest of the paper is organized as follows: Section 2 presents the related works about deep learning models in text classification and class imbalance problems. Section 3 and 4 introduce the proposed and benchmarking methods, respectively. In section 5, experimental results and analysis are provied. The conclusion of this paper is presented in section 6.

\section{Related work}
\label{}

To apply deep learning to solving large-scale text classification problems, the most important task is to obtain text representation. Mikolov \textit{et al}. \cite{mikolov2013distributed}, \cite{mikolov2013efficient} developed word2vec model for computing word vector representation, and the model was well verified in the semantic dimension, which greatly promoted the process of text analysis. However, what word2vec learned is words with similar contexts, and there is still a gap between the learned word vector representation and real semantics. Kim \cite{2014Convolutional} used convolutional neural networks (CNN) originally introduced in computer vision to identify local features of a sentence. Though the TextCNN model achieved good performance in text classification tasks, the hyper-parameter tuning of filters brought some limitations. For example, the improper size of filters could lead to the large computing cost or the loss of long semantic information. If the size of the filter is too large, it will result in computing difficulty. If the size of the filter is too small, it will cause that the model cannot capture long semantic information. Liu \textit{et al}. \cite{2016Recurrent} proposed a multi-task learning model based on the recurrent neural network (RNN). This model utilized the correlation between a related sentence to improve classification performance. Lai \textit{et al}. \cite{lai2015recurrent} developed a recurrent convolutional neural network (RCNN) that combined the recurrent structure and max-pooling layer. This model integrated the advantages of RNN and CNN so that it could learn more contextual information in a sentence. Although CNN and RNN are effective in text classification tasks, they cannot intuitively represent the importance of each sentence and word to the classification category. To overcome this limitation, Zhou \textit{et al}. \cite{2016Attention} added the attention mechanism to Bidirectional Long Short-Term Memory Networks (Att-BLSTM) for capturing the most important semantic information, and it intuitively presented the contribution of each word to the result. Johnson \textit{et al}. \cite{2017Deep} proposed a word-level deep pyramid convolutional neural networks (DPCNN) model to capture the global semantic representation of the text. This model can obtain the best performance by increasing the network depth without increasing too much computational overhead. The state-of-the-art deep learning-based natural language processing model is BERT \cite{devlin2018bert}. Many researchers have studied BERT-based text classification models, offering better performance than previous models in text classification and other natural language processing tasks. As a strong text representation model, BERT can more thoroughly capture the bidirectional semantic relationship in a sentence. Since BERT has learned a good feature representation of text by running a self-supervise learning method on a massive corpus, it is transferred to solve emergency event text classification tasks in this study.

Existing deep learning models in text classification have satisfactory classification performance on balanced data. However, data imbalance can degrade the stability and generalization of these models. Typical data imbalance is data distribution imbalance that reflects different numbers of samples among classes \cite{li2010data}. Consequently, a biased model is yielded based on generic loss functions. The learned model performs well on categories with sufficient samples and performs poorly on classes with fewer samples. In the literature, various methods were developed to solve the problem of data distribution imbalance. One type of methods is sampling methods, including over-sampling, under-sampling, and hybrid sampling. Synthetic minority over-sampling techniques (SMOTE) \cite{chawla2002smote} is one of the most popular over-sampling methods, balancing data distribution via generating samples of minor categories. Raghuwanshi \textit{et al}. \cite{raghuwanshi2020smote} employed SMOTE based class-specific extreme learning machine to increase the classifier's attention to samples of minor categories. Experimental results demonstrated that the algorithm had a high efficacy on real benchmark datasets. Different from the over-sampling, under-sampling randomly samples a subset from  categories with large number of samples. Liu \textit{et al}. \cite{liu2020dealing} utilized random under-sampling to reduce the negative impact of class imbalance. Computational results showed that this method had a better classification performance compared with the classifier without using any data sampling. Based on these types of sampling techniques, some hybrid sampling methods are developed. Li \textit{et al}. \cite{li2020aco} introduced an ant colony optimization resampling method to handle class imbalance problem. This model employed the colony optimization algorithm to get the best subset from the balanced dataset generated via over-sampling. The significant improvement was obtained compared with conventional over-sampling methods. However, sampling methods directly change the original data distribution and two kinds of problems are observed. First, the over-sampling process might introduce extra noises into the dataset, influencing the model training. Second, the under-sampling process might induce information loss of the original data. 

Another strategy to address class imbalance is to increase the loss of the model for misclassified samples. Cao \textit{et al}. \cite{cao2018imcstacking} designed a feature inverse mapping-based cost-sensitive stacking learning model, and it combined cost-sensitive methods with ensemble methods. The effectiveness and efficiency of this method were validated on both linear and ensemble forest classifiers with imbalanced datasets. Shi \textit{et al}. \cite{shi2020penalized} proposed a penalized multiple distribution selection classifier to address imbalanced data problem. The classifier employed a mixture distribution including a softmax distribution and a set of degenerate distributions to fit the imbalanced data. Compared with conventional single softmax distribution, this model had lower computational overhead and higher efficiency in imbalanced data classification tasks.

Inspired by previous works, this study aims to address the limitations of BERT-based models on imbalanced data distribution via adjusting the cost of different samples for emergency text classification. We propose a novel loss function that reduces the cost of correctly classified samples and increases the cost of misclassified samples. Besides, a robust optimizer is employed to update parameters of the BERT-based model. Therefore, the proposed method has better classification performance compared with the generic BERT-based model, and it is feasible to apply the proposed method to emergency events text classification.

\section{The proposed method}
\label{}

In this section, the structure of the proposed BERT-based model is first introduced. Next, the details of the proposed loss function are described. The introduction of the optimization algorithm is presented in the last.

\subsection{BERT}

BERT was firstly proposed by Devlin \textit{et al}. in 2018 \cite{devlin2018bert}. The BERT employed a multi-layer Transformer \cite{vaswani2017attention} structure that reduces the distance between two words in any position via attention mechanism, which effectively solves the long-term dependency problem in natural language processing. Due to BERT considers the context between left and right sides in all layers, it can learn a good feature representation for words through self-supervised learning on a large number of corpora. Therefore, the BERT pre-trained on a large Chinese corpus is employed to extract features of emergency text. The schematic diagram of the proposed method is shown in Figure 1.

\begin{figure}[ht]
	\centering
	\includegraphics[width=1\linewidth]{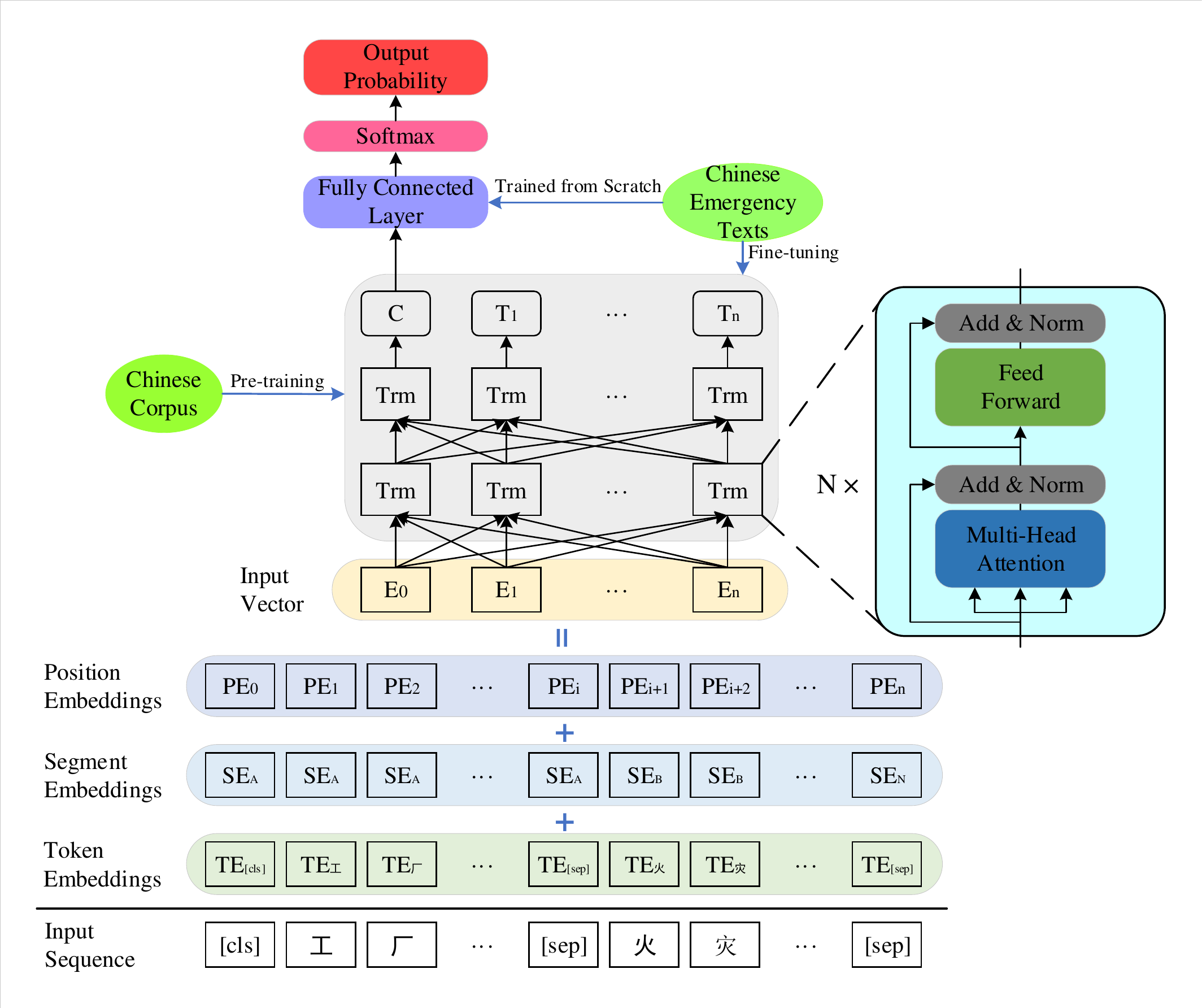}
	\caption{The schematic diagram of the proposed method}
	\label{fig:figure1}
\end{figure}

As shown in Figure 1, the input vector of the BERT-based model is the unit sum of three embedding features that are token embedding, segment embedding, and position embedding. The token embedding is an encoding feature obtained from WordPiece embeddings \textit{et al}. \cite{wu2016google} with a 30,000 token vocabulary. The segment embedding is used to judge the context between two sentences. For two sentences, the segment feature value of the first sentence is zero, and the segment feature value of the second sentence is one. The position embedding is a position feature that represents the position information of each word. Besides, there are two special masked symbols [cls] and [sep]. Between them, the former is a classification model feature, and the latter is used to disconnect two sentences in the input corpus. In the network structure BERT, Trm is an encoder block of the transformer with multi-head attention that can learn multiple representation features from the text. As the main component of the proposed model, BERT is firstly pre-trained on a large Chinese corpus. Based on the pre-trained parameters, the proposed model is fine-tuned on the emergency text dataset. The output probability of the proposed model is calculated in (1) and (2)
\begin{equation}
\vec{y}=WC+\vec{b}
\end{equation}
\begin{equation}
{{p}_{k}}=\frac{{{e}^{{{y}_{k}}}}}{\sum\nolimits_{i=1}^{n}{{{e}^{{{y}_{i}}}}}}
\end{equation}
where $C$ is classification feature vector output by BERT, $W$ and $\vec{b}$ are weights and biases to be trained. Therefore, we can compute a classification loss based on the output probability ${p}_{k}$.

\subsection{The proposed loss function}
Cross-entropy (CE) is a standard classification loss function that is widely used in multi-class classification tasks. It is defined as (3)
\begin{equation}
CE(y,p)=-\sum\limits_{i=1}^{N}{{{y}_{i}}\log ({{p}_{i}})=-\log ({{p}_{k}})}
\end{equation}
where $p$ is the estimated probability distribution, and $y$ is the true probability distribution. For one-hot encoding, only the probability corresponding to the ground truth is 1. 

However, the cross-entropy loss generally does not perform well on imbalanced data.  In the process of network training, the loss of categories with a large number of samples dominates the total training loss, which causes the model to have a bias towards these categories. On the contrary, categories with fewer samples have a relatively low estimated probability due to the small sample size. Considering this limitation, Lin \textit{et al}. \cite{lin2017focal} proposed a focal loss to make the model focus on poorly classified data samples. The focal loss assigns different weights to the samples according to their estimated probability, and it is defined as (4)
\begin{equation}
F(y,p)=-\sum\limits_{i=1}^{N}{{{y}_{i}}{{(1-{{p}_{i}})}^{\gamma }}\log ({{p}_{i}})=-}{{(1-{{p}_{k}})}^{\gamma }}\log ({{p}_{k}})
\end{equation}
where $\gamma$ is a tunable hyper-parameter. When $\gamma = 0$, the focal loss is converted to cross-entropy loss. Compared with cross-entropy, the focal loss has an additional penalty. The penalty increases exponentially as the $\gamma$ increases. Generally, the model can learn sufficient feature information from categories with a large number of samples. Since the model already has a relatively high estimated probability to such samples, the focal loss assigns them a relatively small weight to focus more on poorly classified samples. Besides, it is difficult for the model to accurately classify the categories with a small number of samples due to inadequate training. For these poorly classified samples, the focal loss assigns them a relatively large weight to attract the attention of the model. Therefore, some samples still can provide a large contribution to the total training loss even though their numbers are small. However, focal loss is not enough ideal. When the estimated probability is small, the loss value of the focal loss is greatly weakened compared with that of cross-entropy, especially when the $\gamma$ is large.

Considering the above mentioned limitation, we propose a novel loss function, cross-entropy weighted focal (CEWF) loss, with a tunable weight parameter $t\geq0$, defined as (5)
\begin{equation}
\begin{aligned}
CEWF(y,p) &=-\sum\limits_{i=1}^{N}{{{y}_{i}}\left[ \frac{{{e}^{(1-{{p}_{i}})t}}}{{{e}^{{{p}_{i}}t}}+{{e}^{(1-{{p}_{i}})t}}}\log ({{p}_{i}})+\frac{{{e}^{{{p}_{i}}t}}{{(1-{{p}_{i}})}^{\gamma }}}{{{e}^{{{p}_{i}}t}}+{{e}^{(1-{{p}_{i}})t}}}\log ({{p}_{i}}) \right]} \\ 
& =-\frac{{{e}^{(1-{{p}_{k}})t}}}{{{e}^{{{p}_{k}}t}}+{{e}^{(1-{{p}_{k}})t}}}\log ({{p}_{k}})-\frac{{{e}^{{{p}_{k}}t}}{{(1-{{p}_{k}})}^{\gamma }}}{{{e}^{{{p}_{k}}t}}+{{e}^{(1-{{p}_{k}})t}}}\log ({{p}_{k}})  
\end{aligned}
\end{equation}

\begin{figure}
	\centering
	\includegraphics[width=0.8\linewidth]{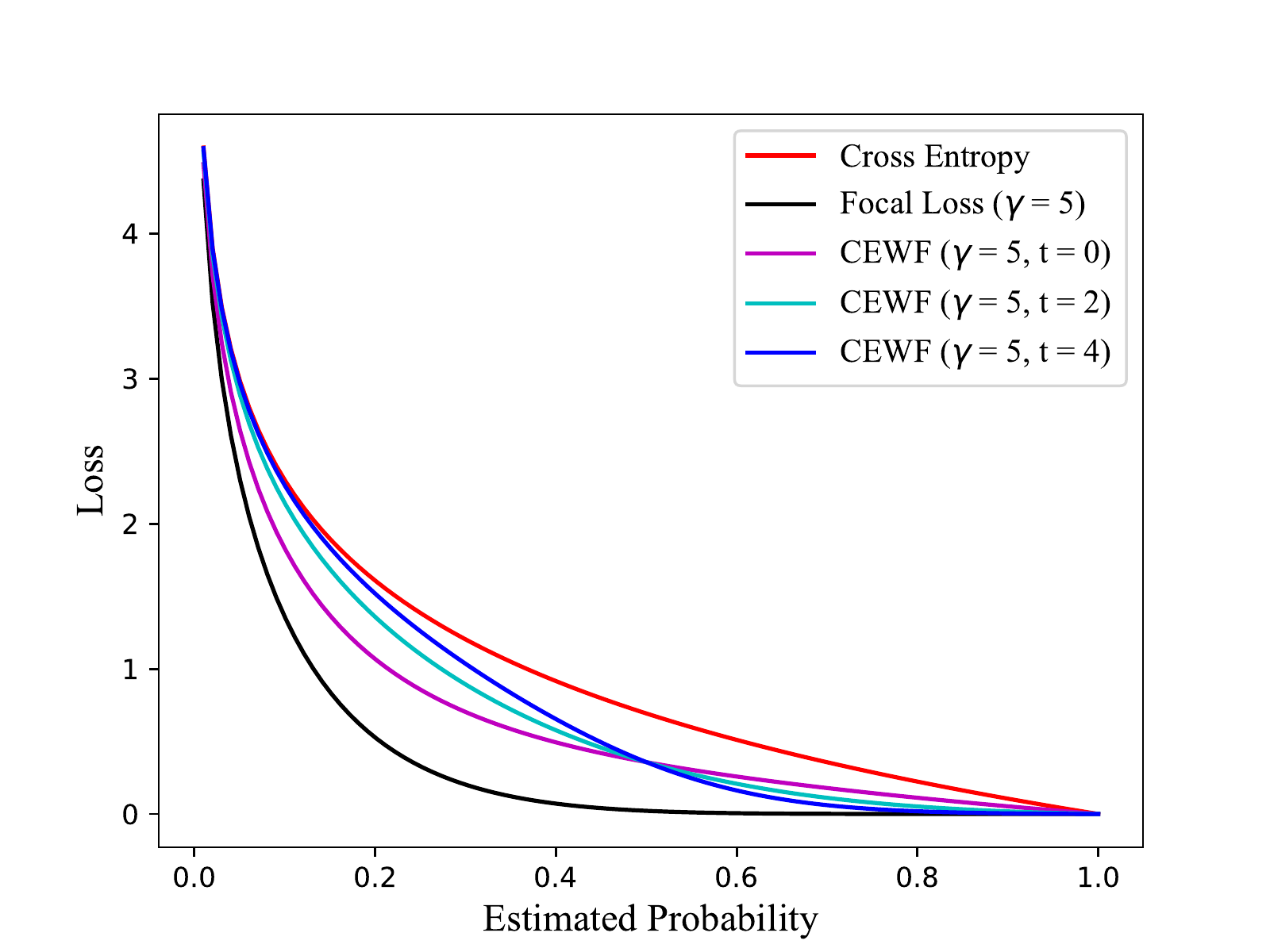}
	\caption{CEWF loss function curves with different $t$}
	\label{fig:tchanges}
\end{figure}

The CEWF loss function curves with different $t$ values are shown in Figure 2. It is observed from Figure 2 that all CEWF loss curves lie between cross-entropy and focal loss. Meanwhile, as the estimated probability increases, the CEWF loss gradually approaches Focal loss. When the estimated probability is small, the CEWF loss is close to cross-entropy. Besides, it can be seen that with larger $t$ values, this trend is more obvious. Therefore, when a sample is classified well, the CEWF loss assigns a loss smaller than the cross-entropy loss to this sample. When a sample is classified poorly, the CEWF loss assigns a loss larger than the focal loss to this sample.

\begin{figure}
	\centering
	\includegraphics[width=0.8\linewidth]{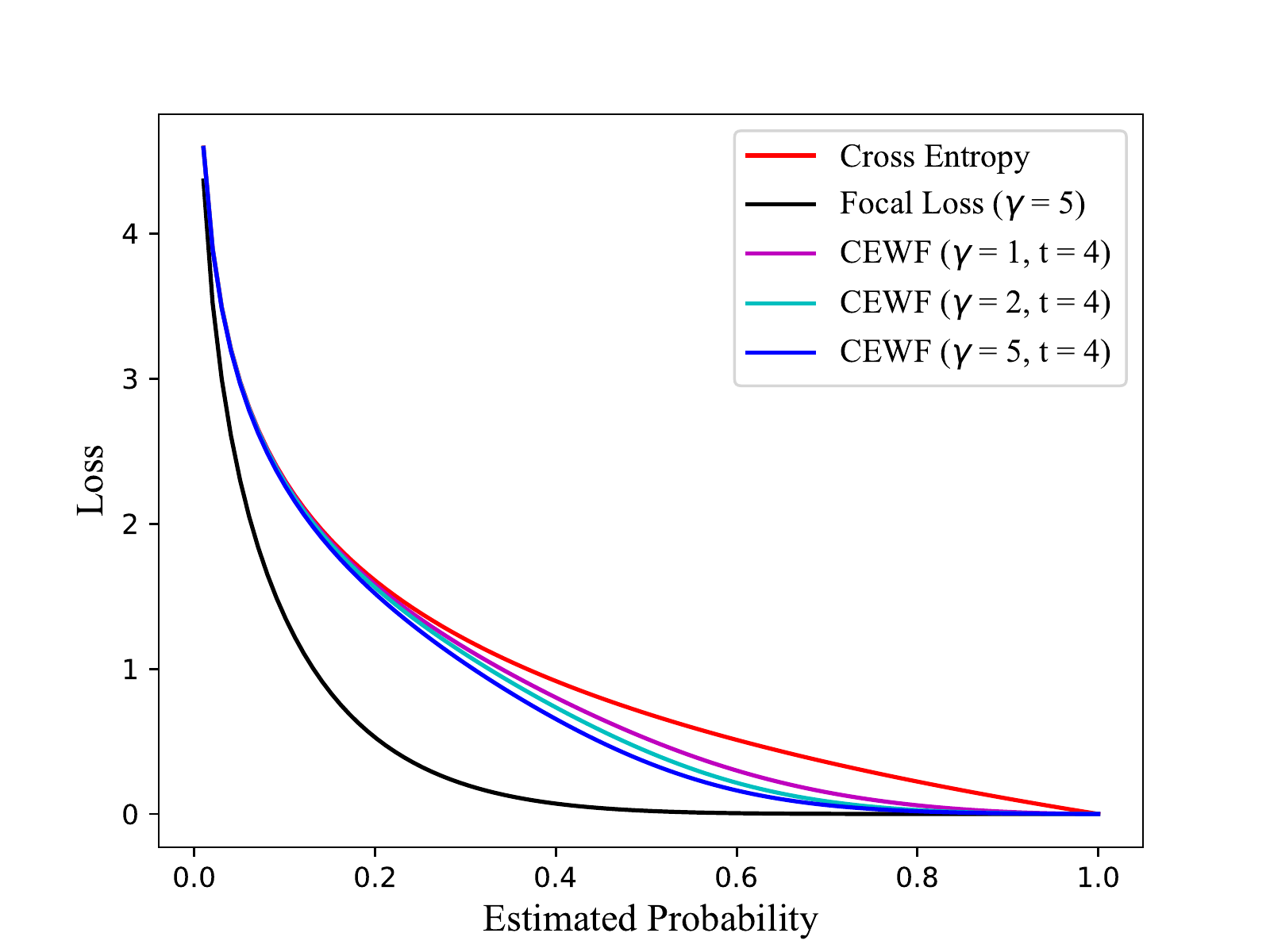}
	\caption{CEWF loss function curves with different $\gamma$}
	\label{fig:gammachanges}
\end{figure}

The CEWF loss function curves with different $\gamma$ values are shown in Figure 3. As shown in Figure 3, $\gamma$ has a little impact on poorly classified samples. When the estimated probability is large, the CEWF loss decreases with the increasing $\gamma$.

In conclusion, the proposed CEWF loss can reduce the difference between the contributions of well classified and poorly classified samples to the total training loss, which can reduce the impact of data distribution imbalance on model training.

\subsection{Optimization algorithm}
Adam is a widely used deep learning optimization algorithm, and it has the advantage of faster convergence than stochastic gradient descent (SGD). However, the learning rate of Adam in the later training stage is small, which affects the effective convergence \cite{keskar2017improving}. Besides, the Adam algorithm may overfit the features that appeared in the early stage, and it is difficult for the features that appear in the later stage to correct the previous fitting effect \cite{wilson2017marginal}. Considering these limitations, Adabound optimization algorithm was proposed by Luo \textit{et al}. \cite{Luo2019AdaBound} in 2019. The Adabound used a dynamic range of learning rate to achieve a gradual smooth transition from Adam to the SGD, which avoids the impact of extreme learning rates. Furthermore, the Adabound has a high learning speed at the beginning of training and a good convergence result at the end of the training. Therefore, the Adabound optimization algorithm is employed to learn the proposed model parameters in this study.

\section{Benchmarking methods}
To assess the performance of the proposed method, five models, BERT with Adabound and CE, the generic BERT classifier (Adam +CE), TextRCNN, Att-BLSTM, and DPCNN, are benchmarked. 

TextRCNN model \cite{lai2015recurrent} combined bi-directional recurrent structure and max-pooling layer was proposed by Lai \textit{et al}. In order to express the meaning of a word more accurately, the author used the word itself and its context to represent the word. In this model, suppose that $c_{l}{(w_{i})}$ is the text on the left of word ${w_{i}}$, and  $c_{r}{(w_{i})}$ is the text on the right of word ${w_{i}}$, they are computed as (6) and (7)
\begin{equation}
{{c}_{l}}({{w}_{i}})=f({{W}^{l}}{{c}_{l}}({{w}_{i-1}})+{{W}^{sl}}e({{w}_{i-1}}))
\end{equation}
\begin{equation}
{{c}_{r}}({{w}_{i}})=f({{W}^{r}}{{c}_{r}}({{w}_{i+1}})+{{W}^{sr}}e({{w}_{i+1}}))
\end{equation}
where $e({{w}_{i-1}})$ is the word embedding of word ${w}_{i-1}$. The recurrent structure can capture all $c_{l}$ in the forward scan of the text and all $c_{r}$ in the reverse scan. The representation of word $w_{i}$ can be defined as (8)
\begin{equation}
{{x}_{i}}=[{{c}_{l}}({{w}_{i}});e({{w}_{i}});{{c}_{r}}({{w}_{i}})]
\end{equation}
Based on the word representation, we can send it to the pooling layer through linear transformation to get the most important information.

Att-BLSTM model \cite{2016Attention} adds the attention mechanism to the BLSTM. In this model, the BLSTM is responsible for capturing information from the left and right sides of a sentence, and the attention mechanism makes the model automatically focus on the words that have a significant impact on classification. Suppose the output vector set of the BLSTM layer is $H$, $[h_{1},h_{2},...,h_{T}]$, where $T$ is the sequence length. The attention is computed in (9)-(11):
\begin{equation}
M=\tanh (H)
\end{equation}
\begin{equation}
\alpha =softmax ({{W}^{T}}M)
\end{equation}
\begin{equation}
r=H{{\alpha }^{T}}
\end{equation}
where $W$ is the parameter that needs to be learned.

DPCNN \cite{2017Deep} is a deep pyramid CNN model which can effectively capture the long-term dependence in the text. After converting the discrete text information into continuous representation, DPCNN simply stacks the convolution module and the downsampling layer, which makes the model have a small computational overhead. Meanwhile, an additive shortcut connection with identity mapping was used in this model, computed as $z+f(z)$ where $f(z)$ is the convolutional layer which is short-circuited. In DPCNN, the pre-activation was employed when calculating $f(z)$, and it is defined as (12)
\begin{equation}
f(z)=W\sigma (z)+b
\end{equation}
where both $W$ and $b$ are trainable parameters, $z$ is text region embedding, and $\sigma$ is $sigmoid$ activation function.

\section{Case study}
To validate the feasibility of the proposed method, the proposed method is tested on a real Chinese emergency text dataset, and its results are compared with five benchmarking methods.
\subsection{Data description}
In this study, an emergency text dataset in Chinese containing 9649 samples was collected from an emergency management website (http://www.safehoo.com), and it is manually labeled by experts in the field of emergency management. In this dataset, 8 common types of emergency accidents are included, such as air crash, fire, traffic accident, \textit{etc}. A detailed description of the dataset is presented in Table 1. It can be seen from Table 1 that there is a large difference in the sample size of different emergency accidents. In particular, the ratio of the number of fire samples to that of air crash samples nearly reaches 141:1. It can significantly reflect the robustness of all considered methods against imbalanced data distribution.

\begin{table}
	\caption{Emergency event text dataset}
	\centering
	\begin{tabular}{lll}
		\toprule
		Emergency accident	  &Amount of data      &Label  \\
		\midrule
		Air crash            &19         &0       \\
		Electric shock       &1726       &1       \\
		Fall				 &1323       &2       \\
		Fire				 &2671       &3       \\
		Scald				 &378        &4       \\
		Crane accident 	     &689        &5       \\
		Struck by objects 	 &612        &6       \\
		Traffic accident     &2231       &7       \\
		\bottomrule
	\end{tabular}
\end{table}

\subsection{Assessment metrics}

To assess the classification performance of different methods, the confusion matrix is considered. Based on the confusion matrix, four metrics, Accuracy, Weighted-Precision (WPrecision), Weighted-Recall (WRecall), Weighted-F1 (WF1), are computed according to (13)-(16):
\begin{equation}
	\text{Accuracy}=\frac{1}{N}\sum\nolimits_{i=1}^{m}{T{{P}_{i}}}
\end{equation}
\begin{equation}
	\text{WPrecision} =\frac{1}{N}\sum\nolimits_{i=1}^{m}{{{n}_{i}}\frac{T{{P}_{i}}}{T{{P}_{i}}+F{{P}_{i}}}}
\end{equation}
\begin{equation}
	\text{WRecall} =\frac{1}{N}\sum\nolimits_{i=1}^{m}{{{n}_{i}}\frac{T{{P}_{i}}}{T{{P}_{i}}+F{{N}_{i}}}}
\end{equation}
\begin{equation}
	\text{WF1} =\frac{1}{N}\sum\nolimits_{i=1}^{m}{{{n}_{i}}\frac{2{{P}_{i}}{{R}_{i}}}{{{P}_{i}}+{{R}_{i}}}}
\end{equation}
where $N$ is the total number of samples, $m$ is the total number of classes, ${n}_{i}$ is the number of samples of the $i$-th class, $TP_{i}$ is the number of samples correctly classified by the model in the i-th class, $FP_{i}$ is the number of negative samples misclassified by the model in the i-th class, $FN_{i}$ is the number of positive samples misclassified by the model in the $i$-th class. $P_{i}$ and $R_{i}$ are the precision and recall of the $i$-th class, respectively.

\subsection{Experiment and analysis}
To assess the generalization performance of the proposed model, all data samples are divided into training set, validation set, and test set, and the division ratio is 6:2:2. All considered algorithms are implemented on a workstation with AMD Ryzen 9 3950x@2.2 GHz CPU and 32GB RAM, as well as a Nvidia RTX 3080 GPU with 10GB memory. The program based on Python 3.7 is executed on Ubuntu 20.04. As the main framework of the proposed method, BERT is pre-trained on a large Chinese corpus. Based on the pre-trained parameters, the proposed model is fine-tuned on the emergency text dataset, where the batch size is set to 8, the learning rate is set to $5\times {{10}^{-5}}$, and each text sample is cropped into a sequence of length 400. The accuracy of all considering methods on the validation set is shown in Figure 4.
\begin{figure}
	\centering
	\includegraphics[width=0.7\linewidth]{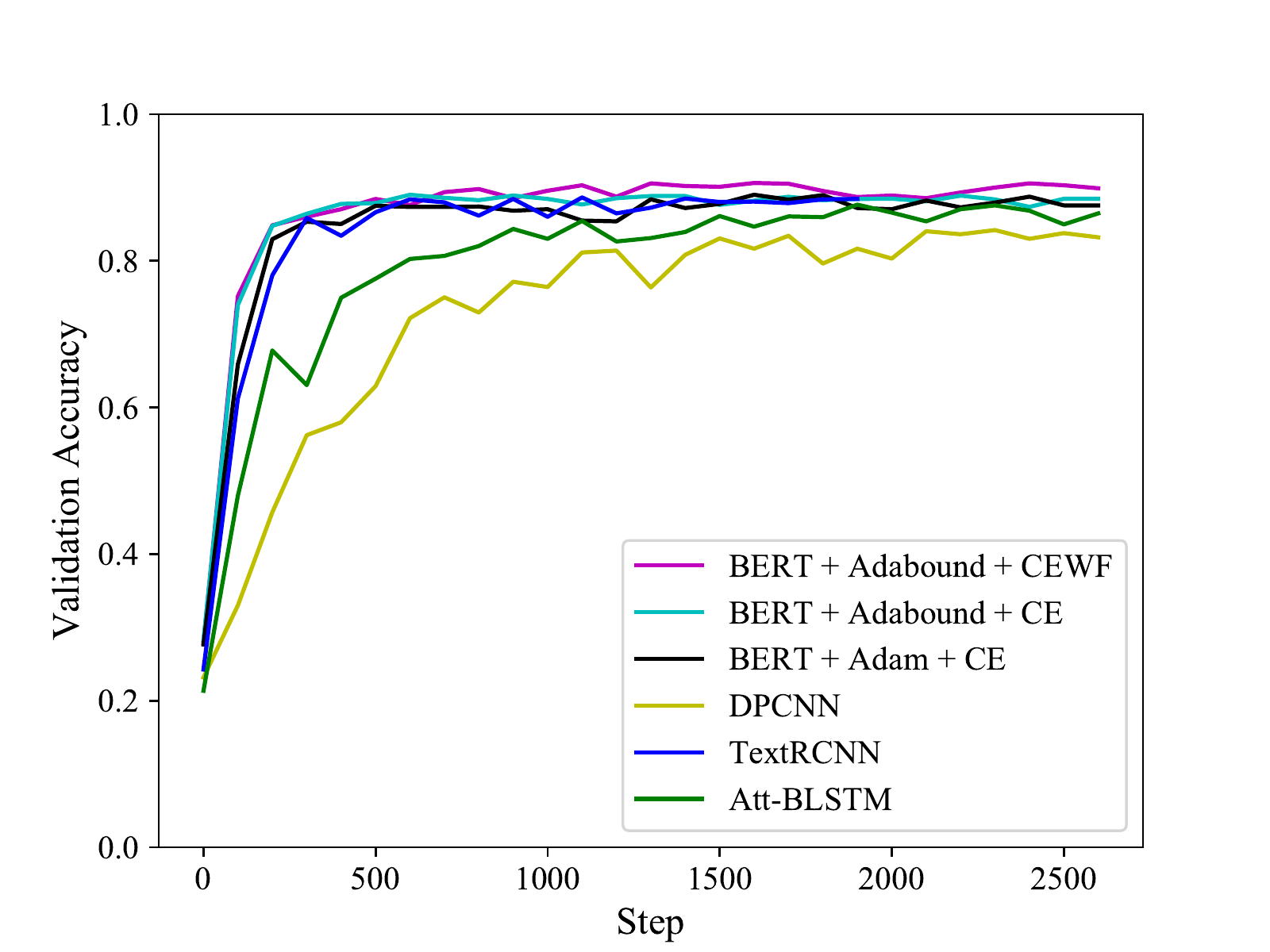}
	\caption{The accuracy of all considering methods on the validation set}
	\label{fig:validation-accuracy}
\end{figure}

As shown in Figure 4, The proposed method converges to the highest validation accuracy among all considering methods. Meanwhile, under the same loss function, Adabound has a better-converged result compared with Adam. Based on Adabound optimization, the proposed loss function further improves the accuracy of the classification model.

\begin{figure}
	\centering
	\includegraphics[width=0.7\linewidth]{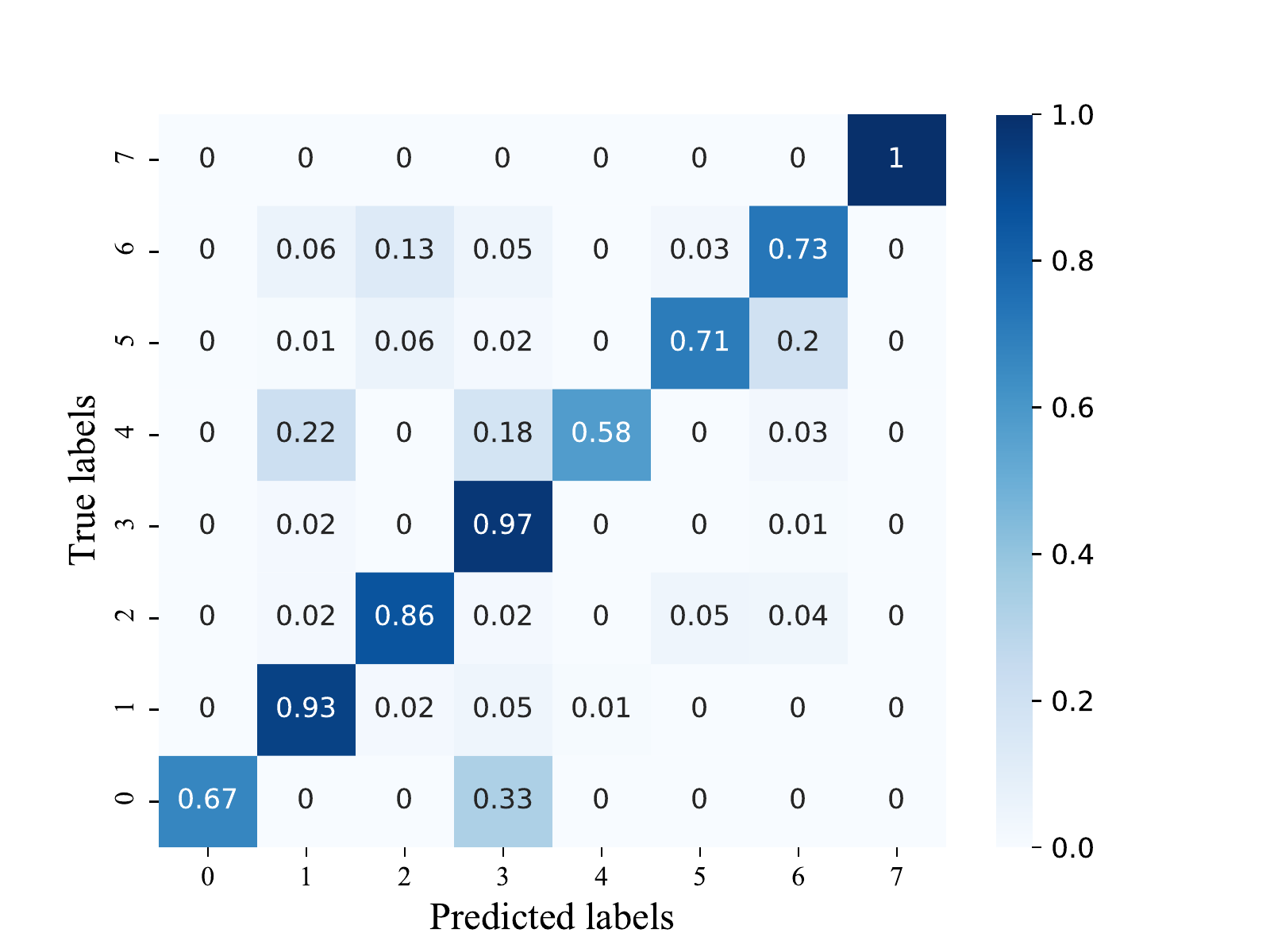}
	\caption{Confusion matrix of the generic BERT classifier}
	\label{fig:adamce}
\end{figure}
\begin{figure}
	\centering
	\includegraphics[width=0.7\linewidth]{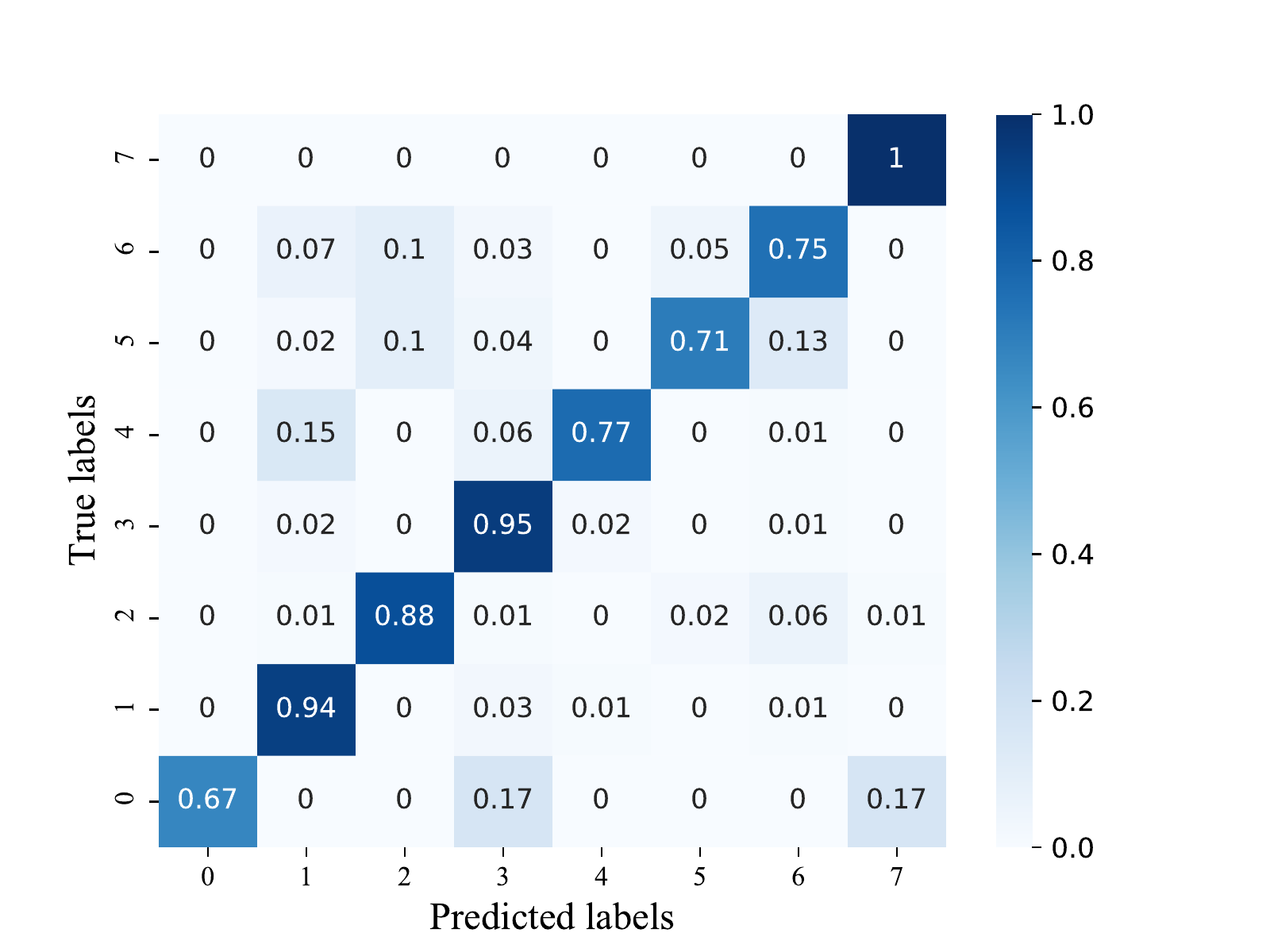}
	\caption{Confusion matrix of the proposed method}
	\label{fig:proposed}
\end{figure}

To further illustrate the performance of the proposed method, the normalized confusion matrices of the generic BERT-based classification model and the proposed method are shown in Figures 5 and 6. From Figures 5 and 6, it can be seen that the proposed method has more steady classification performance compared with the generic BERT-based classifier over different classes. In particular, the classification accuracy for samples of the fourth category (sample size: 378) has been improved from 0.58 to 0.77. To evaluate the performance of the model comprehensively, the assessment metrics of different models on the test set are computed and presented in Table 2.
\begin{table}
	\caption{Computational results of different models on test set}
	\centering
	\begin{tabular}{lllll}
		\toprule
		Model	                              &Accuracy               &WPrecision          &WRecall        &WF1  \\
		\midrule
		BERT + Adabound + CEWF                 &0.9161	              &0.9174	           &0.9161	       &0.9155       \\
		BERT + Adabound + CE                   &0.9067	              &0.9132	           &0.9067	       &0.9072       \\
		BERT + Adam + CE			           &0.9047	              &0.9067	           &0.9047	       &0.9029       \\
		TextRCNN				               &0.8964	              &0.9059	           &0.8964	       &0.8983       \\
		DPCNN				                   &0.8705	              &0.8669	           &0.8705	       &0.8666       \\
		Att-BLSTM 	                           &0.8684	              &0.8728	           &0.8684	       &0.8669       \\
		\bottomrule
	\end{tabular}
\end{table}
According to the results presented in Table 2, it is obvious that the proposed method dominates all benchmarks in terms of the highest accuracy, WPrecision, WRecall, and WF1 values. In addition, the performance of the BERT-based method is generally better than conventional neural network methods. Compared with Adam, Adabound improves the weighted precision of the BERT-based model from 0.9067 to 0.9132. Furthermore, a significant performance improvement is yielded by using the proposed CEWF loss function compared with the cross-entropy in the generic BERT-based model. Therefore, it is promising to apply the proposed method to solving Chinese text classification problem for emergency domain.

\section{Conclusion}

This paper proposed an automatic Chinese text classification method for emergency domain. In the proposed method, a novel loss function, the CEWF loss function, was proposed to improve the performance of the BERT-based model on the imbalanced dataset. To avoid impacts of the extreme learning rate, the Adabound optimization algorithm that achieved a gradual smooth transition from Adam to SGD was employed to learn parameters of the proposed model. The feasibility and effectiveness of the proposed method were validated on the real Chinese emergency text dataset. Meanwhile, the proposed method was compared with other benchmarking methods. 

Experimental results showed that the proposed loss function can effectively address the problem of insufficient training caused by fewer samples from minor classes, and thus the performance of the BERT-based model was significantly improved over different classes. Furthermore, the Adabound optimizer better tuned the model compared with Adam optimizer. Therefore, it is feasible to apply the proposed algorithm in smart emergency management systems.





\bibliographystyle{elsarticle-num} 
\bibliography{references} 
\end{document}